\documentclass[letterpaper]{article} 
\usepackage[draft]{aaai25}  
\usepackage{times}  
\usepackage{helvet}  
\usepackage{courier}  
\usepackage[hyphens]{url}  
\usepackage{graphicx} 
\usepackage{tikz}
\usepackage{amsmath}
\usepackage{tabularx}
\usepackage{natbib}  
\usepackage{caption} 
\usepackage{algorithm}
\usepackage{algorithmic}
\usepackage{booktabs}
\usepackage{url}            

\usepackage{newfloat}
\usepackage{listings}
\DeclareCaptionStyle{ruled}{labelfont=normalfont,labelsep=colon,strut=off} 
\floatstyle{ruled}
\newfloat{listing}{tb}{lst}{}
\floatname{listing}{Listing}
\title{HiRED: Attention-Guided Token Dropping for Efficient Inference of High-Resolution Vision-Language Models}

\author{
    Kazi Hasan Ibn Arif\textsuperscript{\rm 1}, JinYi Yoon\textsuperscript{\rm 1}, Dimitrios S. Nikolopoulos\textsuperscript{\rm 1}, Hans Vandierendonck\textsuperscript{\rm 2},
    \\Deepu John\textsuperscript{\rm 3}, Bo Ji\textsuperscript{\rm 1}
}
\affiliations{
    \textsuperscript{\rm 1}Virginia Tech, Blacksburg, VA, USA\\
    \textsuperscript{\rm 2}Queen’s University Belfast, Belfast, UK\\
    \textsuperscript{\rm 3}University College Dublin, Dublin, Ireland\\

}

\usepackage{bibentry}
\usepackage{xcolor}         
\usepackage{graphicx}
\usepackage{subcaption}
\usepackage{multirow}
\usepackage{algorithm}
\usepackage{algorithmic}
\usepackage{sidecap}
\usepackage{color}
\usepackage{colortbl}
\usepackage{diagbox}
\usepackage{amssymb}
\usepackage{pifont}
\newcommand{\xmark}{\ding{55}}%
\usepackage{arydshln}
\usepackage{amsmath}

\newcommand{\refsec}[1]{Section\@~\ref{sec:#1}}
\newcommand{\reffig}[1]{Fig\@.~\ref{fig:#1}}

\newtheorem{insight}{Insight}

\definecolor{LightOrange}{rgb}{1.0, 0.9, 0.8}
\definecolor{DeeperOrange}{rgb}{1.0, 0.85, 0.7}
\definecolor{LightGreen}{rgb}{0.8, 1.0, 0.8}

\newcommand{\algoname}{HiRED}

\begin{document}

\maketitle

\begin{abstract}

High-resolution Vision-Language Models (VLMs) are widely used in multimodal tasks to enhance accuracy by preserving detailed image information. However, these models often generate an excessive number of visual tokens due to the need to encode multiple partitions of a high-resolution image input. Processing such a large number of visual tokens through multiple transformer networks poses significant computational challenges, particularly for resource-constrained commodity GPUs. To address this challenge, we propose \emph{High-Resolution Early Dropping (HiRED)}, a plug-and-play token-dropping method designed to operate within a fixed token budget. HiRED leverages the attention of \texttt{CLS} token in the vision transformer (ViT) to assess the visual content of the image partitions and allocate an optimal token budget for each partition accordingly. The most informative visual tokens from each partition within the allocated budget are then selected and passed to the subsequent Large Language Model (LLM). We showed that HiRED achieves superior accuracy and performance, compared to existing token-dropping methods. Empirically, HiRED-20\% (i.e., a 20\% token budget) on LLaVA-Next-7B achieves a 4.7$\times$ increase in token generation throughput, reduces response latency by 78\%, and saves 14\% of GPU memory for single inference on an NVIDIA TESLA P40 (24~GB). For larger batch sizes (e.g., 4), HiRED-20\% prevents out-of-memory errors by cutting memory usage by 30\%, while preserving throughput and latency benefits.

\end{abstract}

%
\begin{links}
    \link{Code}{https://github.com/hasanar1f/HiRED}
\end{links}

\section{Introduction}\label{sec:introduction}

Vision-Language Models (VLMs), such as GPT-4v~\cite{gpt4vtechnicalreport}, Gemini~Pro~\cite{gemini-1.5-pro}, LLaVA~\cite{llava-main}, and Qwen-VL~\cite{qwen-vl-main}, have emerged as remarkable multimodal models that learn from visual and textual data. However, these VLMs inherently work for low-resolution images only and would lose fine-grained visual information if applied to high-resolution images~\cite{survey-on-vlm-mm-llms,internlm-xcomposer2}. 
To address this issue, recent VLMs, referred to as high-resolution VLMs, employ dynamic partitioning to encode high-resolution images~\cite{llava-next-technical,internlm-xcomposer2, monkey-main, text-monkey-main, sphinx-main, llava-med-high-res}. 

\begin{figure}[!t]
    \centering
    \includegraphics[width=0.95\linewidth]{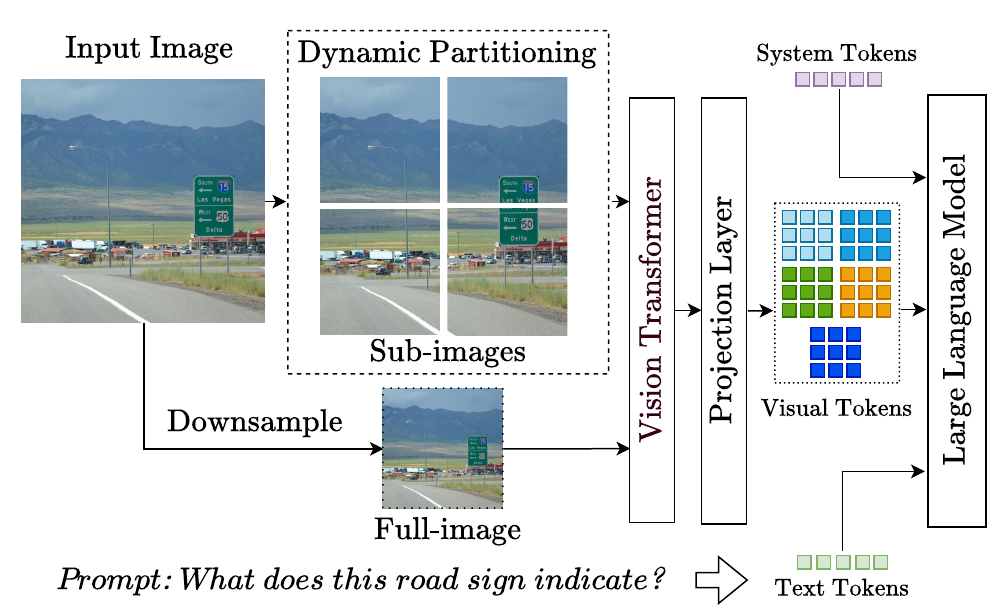}
    \caption{Inference steps of LLaVA-Next~\cite{llava-next-technical} for a high-resolution VLM with dynamic partitioning.
    }
    \label{fig:inference-step-of-llava-next}
\end{figure}

A typical inference pipeline of high-resolution VLMs with dynamic partitioning is illustrated in \reffig{inference-step-of-llava-next}. Specifically, a high-resolution input image is partitioned into multiple sub-images (e.g., four sub-images for a square image in LLaVA-Next); a downsampled version of the original image, referred to as the full-image, is also included. Subsequently, a vision encoder such as Vision Transformers (ViTs) encodes each low-resolution image partition into image features, which are then converted to visual tokens in the text embedding space through a lightweight Projection Layer. These visual tokens are concatenated and fed into a Large Language Model (LLM) (along with text tokens and system tokens) to generate the final response. Here, the full-image and the sub-images (commonly referred to as image partitions in this paper) have different amounts of visual content and thus, exhibit different degrees of importance. While the full-image captures the global context of the original image, each sub-image is for a more detailed local representation of corresponding specific areas. This multi-partitioning approach enables the inclusion of more visual details, which can significantly boost accuracy. For example, accuracy can be improved by 15\% when the image resolution is increased from 336$\times$336 to 1344$\times$1344~\cite{mm1-apple-main}.

\begin{table*}[!t]
    \centering
    \caption{Comparison between our \algoname{} and existing methods.
    } 
    \label{tab:compare-vlm}
    \begin{tabular}{lccccc}
        \toprule
          Method & High Resolution & Token Budget & Early Dropping & Task Coverage  \\\midrule
        FastV~\cite{fastv}  & \xmark &  \checkmark & \xmark & \checkmark \\
        FlexAttention~\cite{flexattention} & \checkmark & \xmark & \xmark  & \checkmark  \\
        TokenCorrCompressor~\cite{token-corr-compress} & \checkmark & \xmark & \checkmark & \xmark \\
        PruMerge~\cite{llava-prumerge}  & \xmark &  \xmark & \checkmark & \checkmark  \\ 
        \midrule
        \textbf{\algoname{} (Ours)} & \checkmark & \checkmark & \checkmark & \checkmark \\
        \bottomrule
    \end{tabular}
\end{table*}

However, due to the need to encode multiple image partitions, high-resolution VLMs often generate 3-10$\times$ more visual tokens than their low-resolution counterparts~\cite{internlm-xcomposer2,docowl-1.5}. Such excessive visual tokens result in lower inference throughput, increased generation latency, and higher GPU memory usage. Furthermore, depending on downstream tasks, the number of visual tokens required to represent an image also varies significantly~\cite{matryoshka-mm}. However, most commodity GPUs, such as the Jetson Orin NX (8 or 16 GB) and NVIDIA Tesla T4 (16 GB), have limited computational cores and memory. The quadratic complexity of transformers~\cite{transformers} makes it challenging to process a large number of tokens on these GPUs. In addition, increased key-value (KV) cache size due to storing token embeddings at runtime could cause out-of-memory issues. Therefore, controlling and optimizing the number of visual tokens is essential to meet the system resource constraints. Although traditional optimization techniques (e.g., model quantization, weight pruning, and lightweight architectures) can reduce model size, they do not address the critical issue of excessive visual tokens.

\emph{We aim to achieve efficient inference of high-resolution VLMs 
through strategic dropping of excessive visual tokens.} Such token-dropping schemes are expected to offer four desired properties: 
(i)~\emph{Supporting high-resolution}: plug-and-play integration (i.e., without model training and architectural changes) that promotes easy adoption with existing high-resolution VLMs while maintaining superior accuracy;
(ii)~\emph{Controlling token budget}: having control over the number of visual tokens fed into the LLM to enable efficient inference under various resource constraints and task requirements;
(iii)~\emph{Facilitating early dropping}: dropping tokens in the image encoding stage (i.e., before the generation phase using LLM) 
to reduce input length and enhance computational efficiency; and
(iv)~\emph{Wide task coverage}: covering a wide range of vision-language tasks (vision question answering, image captioning, document understanding, etc.).

The recent few months have witnessed exciting progress towards the above goals~\cite{fastv,flexattention,token-corr-compress,llava-prumerge}. However, none of these works achieve all the aforementioned properties, which are highly desired for efficient high-resolution VLM inference 
(see Table~\ref{tab:compare-vlm} for a summary and Section~\ref{sec:related_work} for a detailed discussion).

\paragraph{Contributions.}
Our work bridges this critical gap and makes the following main contributions:

We propose \emph{High-Resolution Early Dropping (\algoname{})}, a plug-and-play token-dropping framework for efficient inference of high-resolution VLMs. \algoname{} enables attention-guided early dropping of visual tokens under resource constraints and covers a wide range of multimodal tasks. \emph{To the best of our knowledge, \algoname{} is the first framework that achieves all of these desired properties.}

To realize \algoname{}, our key design leverages two crucial insights from attention patterns in ViT. \emph{First}, class token (\texttt{CLS}) to patch token attentions from initial ViT layers are closely correlated with the visual contents and can be used to identify the main objects and irrelevant backgrounds in an image. To allocate a larger budget to a partition with more content, we introduce the \emph{visual content score} (which represents the amount of visual content a partition carries) as the token budget for each sub-image. Second, we observe that \texttt{CLS}-attention (attention scores between the class token and patch tokens of ViT) from the final layers indicate the informativeness of patch tokens. Therefore, we use the \texttt{CLS}-attention (aggregated across multiple heads) from the final layer as \emph{feature importance score} and select tokens with the highest feature importance score within the allocated budget. \emph{By leveraging such \texttt{CLS}-attention patterns in ViT, we design a lightweight yet efficient algorithm for budget allocation and token dropping, two key components of \algoname{}.}

Finally, we implement \algoname{} on three popular open-source VLMs: LLaVA-Next~\cite{llava-next-technical}, LLaVA~\cite{llava-main}, and ShareGPT4V~\cite{sharegpt} and evaluate the accuracy for eight tasks. Our experimental results show that \algoname{}-20\% (i.e., the budget is set to 20\% of the total number of tokens) on LLaVA-Next-7B (a high-resolution VLM) achieves a 4.7$\times$ increase in token generation throughput (2.30 vs. 0.49 tokens/sec), reduces response latency by 78\% (4.21 vs. 19.49 seconds), and saves 14\% of GPU memory (13.76 vs. 16.04 GB) for single inference on an NVIDIA TESLA P40 GPU. For larger batch sizes (e.g., 4), where the 24~GB GPU encounters out-of-memory (OOM) errors with the full token budget, \algoname{}-20\% reduces memory usage by 30\% (16.99 GB) while maintaining the throughput and latency improvements. Moreover, \algoname{} achieves significantly higher accuracy than previous early-dropping methods, such as PruMerge and PruMerge+~\cite{llava-prumerge} across various tasks.

\section{Related Work}\label{sec:related_work}

We categorize highly related works into three groups.

\paragraph{Lightweight Architectures.} Traditional methods often aim to downsize VLMs by reducing the model size, such as LLaVA-Phi-2.7B~\cite{llava-phi}, TinyLLaVA-3.1B~\cite{tiny-llava}, and MobileVLM-3B~\cite{mobile-vlm}. However, these approaches significantly compromise reasoning capabilities due to the substantial reduction in model parameters. Techniques like model quantization~\cite{int8-quantization} and weight pruning or masking~\cite{pruning-wanda} further reduce resource demands but fail to address the critical issue of excessive visual tokens. Token ensemble frameworks such as CrossGET~\cite{crossget} and MADTP~\citet{madtp} primarily target cross-modal transformer-based VLMs, such as BLIP~\cite{blip2}. Methods like Q-Former, M$^3$~\cite{matryoshka-mm}, and Abstractor~\cite{abstractor} require expensive training and fine-tuning.

\paragraph{Sparse Attention Computation in LLM and ViT.} These methods aim to reduce the computational cost of attention mechanisms in the transformer layers
. FastV~\cite{fastv} identifies important visual tokens in the initial layers of LLM and skips unimportant tokens in subsequent layers, but it is not designed for high-resolution VLMs. While FlexAttention~\cite{flexattention} can handle high-resolution images, it does not allow control over the number of visual tokens based on resource constraints. Methods such as DynamicViT~\cite{dynamicvit}, PuMer~\cite{pumer}, and EViT~\cite{evit} are primarily for ViTs, and thus, their efficiency gains are limited as the majority of computation occurs in the LLM.

\paragraph{Early Dropping of Visual Tokens.} Methods like TokenCorrCompressor~\cite{token-corr-compress} and PruMerge~\cite{llava-prumerge} drop visual tokens from the image encoding stage before feeding them to the LLM for greater efficiency. TokenCorrCompressor identifies repetitive whitespace patterns in document images through token-to-token cosine similarity and drops redundant tokens with high similarity. However, their work considers document understanding tasks only. PruMerge prunes out visual tokens with low \texttt{CLS}-attention and merges them with selected tokens, and PruMerge+, an enhanced version, selects additional spatial tokens to mitigate information loss. Since PruMerge is designed for low-resolution VLMs, the model accuracy degrades significantly for high-resolution VLMs (see \refsec{accuracy_evaluation}). Moreover, these methods lack control over the number of visual tokens within the memory budget, which is essential in resource-constrained environments.

\section{Key Insights}\label{sec:key_insights}

\begin{figure}[!t]
    \centering
    \begin{subfigure}[t]{0.48\columnwidth}
        \centering
        \includegraphics[width=\linewidth]{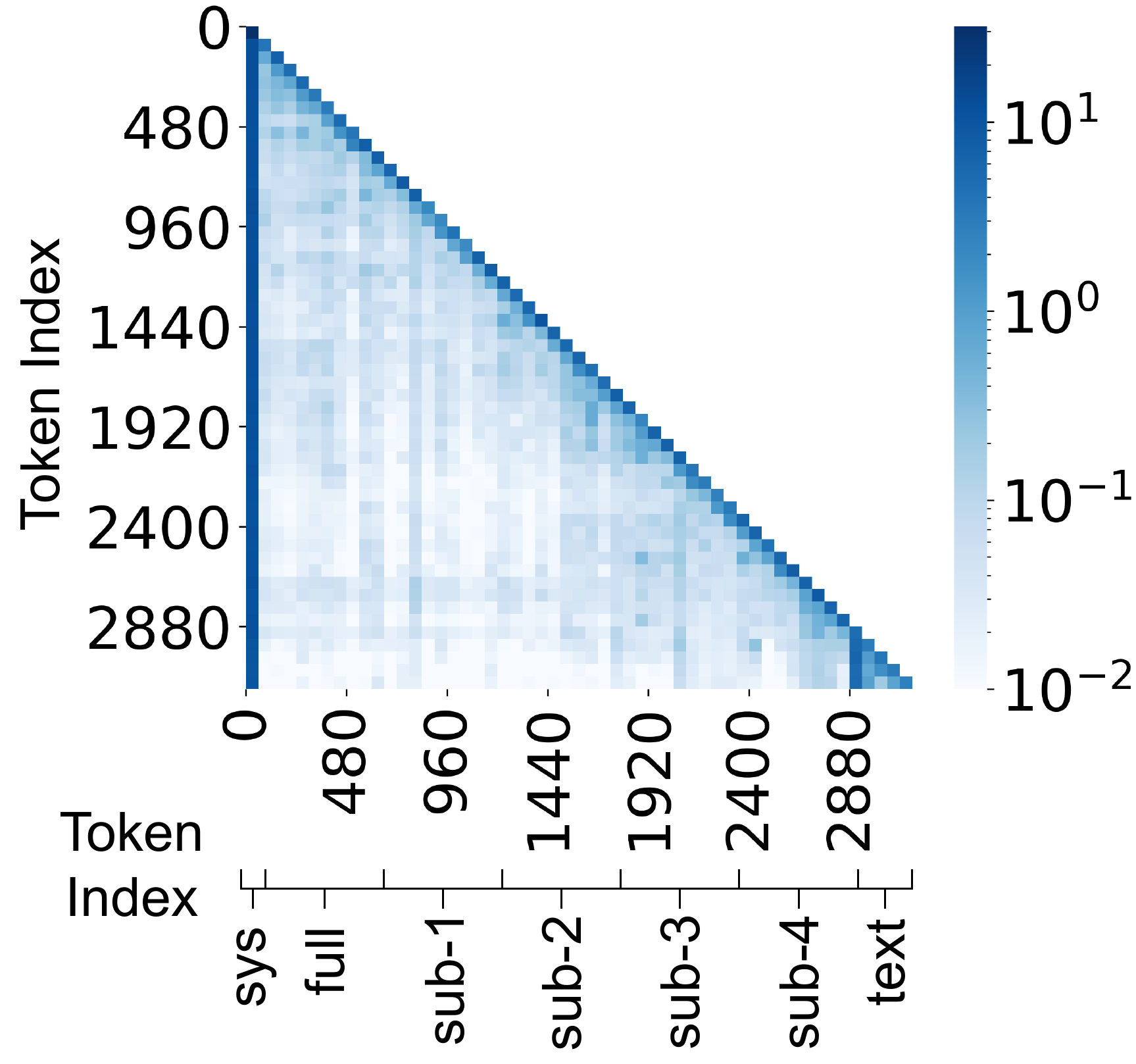}
        \caption{Attention scores of tokens}
        \label{fig:llm_attn_map}
    \end{subfigure}
    \begin{subfigure}[t]{0.48\columnwidth}
        \centering
        \includegraphics[width=1\linewidth]{figure/cumulative_llm_attn.pdf}
        \caption{CDF of attention scores}
        \label{fig:cumulative_llm_attn}
    \end{subfigure}
    \caption{The sparse nature of visual tokens is evident during the generation using LLM. (a) Visual tokens receive significantly less attention compared to system and text tokens. (b) The top 20\% and 40\% of visual tokens account for 60\% and 80\% of the total attention, respectively.}
    \label{fig:sparsity_of_vision_tokens}
\end{figure}

\begin{table}[!t]
    \centering
    \caption{Distribution of the top (10\% and 20\%) visual tokens for the image partitions shown in \reffig{inference-step-of-llava-next}. Here, \textit{TL}, \textit{TR}, \textit{BL}, and \textit{BR} represent the top-left, top-right, bottom-left, and bottom-right corners of the image partitions, respectively.}

    \setlength{\tabcolsep}{4pt}
    \label{table:count_tokens_in_images}
    \begin{tabular}{lccccc}
        \toprule
         Budget & \textit{Full} & \textit{TL} & \textit{TR} & \textit{BL} & \textit{BR} \\
        \midrule
        10\% (= 288~tokens) & 37 & 12 & 29 & 84 & 126 \\
        20\% (= 576~tokens) & 94 & 40 & 61 & 148 & 234 \\
        \bottomrule
    \end{tabular}
\end{table}

\paragraph{Sparse visual tokens with high attention scores.}

In a typical VLM, the LLM processes visual, text, and system tokens together. To understand the role of various tokens in the LLM generation phase, we investigate their attention patterns on LLaVA-Next-7B (see \reffig{llm_attn_map}). This experiment reveals that while visual tokens amount to 80-90\% of all the tokens, they receive significantly less attention than system and text tokens. To further examine the gap between tokens with high and low attention scores, we compute the Cumulative Distribution Function (CDF) of attention scores for top visual tokens with the highest attention scores (see \reffig{cumulative_llm_attn}). The results reveal that a small subset of visual tokens brings most of the context from the image to the LLM.

\begin{insight}[Visual token sparsity]\label{insight:token-sparsity}
    Despite the large number of visual tokens, only a small subset is important in the LLM generation phase, suggesting an opportunity to drop less important tokens without sacrificing accuracy.
\end{insight}

\paragraph{Various Importance of Sub-images.}
In Section~\ref{sec:introduction}, we discussed how dynamic partitioning can significantly boost accuracy by encoding global and detailed local representations through full-image and sub-images. To understand the contribution of image partitions in the LLM generation phase, we count the number of tokens with the highest attention for each image partition in Table~\ref{table:count_tokens_in_images}. It shows that the distribution of top visual tokens varies across different image partitions.

\begin{insight}[Sub-images with different content amounts]\label{insight:diff-budget}
    The variation in the visual content weights of image partitions suggests that some partitions may allow more token dropping than others.
\end{insight}

\section{Our Design: \algoname}

The above observations suggest that only a subset of visual tokens is crucial during the LLM generation phase and the varying importance of image partitions presents a clear opportunity for dropping various numbers of visual tokens from image partitions under a fixed token budget. Motivated by these useful insights, we further explore the \texttt{CLS}-attention pattern in ViT (\refsec{vit_attn_guided_dropping}) and propose a novel \emph{attention-guided token-dropping} scheme (\refsec{design}).

\subsection{\texttt{CLS}-attention Pattern in ViT} \label{sec:vit_attn_guided_dropping}

As discussed in \refsec{key_insights}, it is straightforward to identify important visual tokens (those with higher attention scores) during the LLM generation phase. However, to reduce input sequence length and improve computational efficiency, it is crucial to identify important tokens earlier, before the generation phase. To do this, we utilize the \texttt{CLS}-attention of ViT. Particularly, CLIP~\cite{clipvit} 
splits an input image into a fixed number of non-overlapping patches. In the transformer layers, the \texttt{CLS} token is employed to extract useful information from the patch tokens. By design, the \texttt{CLS}-attention indicates the importance of patches \cite{interpreting-clip}. Since the patch tokens are later transformed into visual tokens for the subsequent LLM, a guided selection of patches can effectively reduce the number of tokens. Now, a key question is: \emph{How do we identify important visual tokens before the LLM generation phase that are crucial for generating the response?}

\begin{figure}
    \centering
    \includegraphics[width=\linewidth]{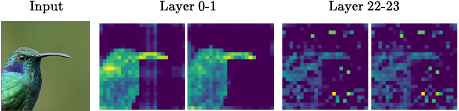}
    \caption{In ViT, \texttt{CLS}-attention map shows distinct characteristics across layers
    . The initial layers highlight the subject patches while ignoring the background, aligning mostly with the image content. The final layers, however, highlight informative patches where ViT stores most of the image features.}
    \label{fig:vit-cls-attn}
\end{figure}

\begin{figure}[t!]
    \centering
    \includegraphics[width=1\linewidth]{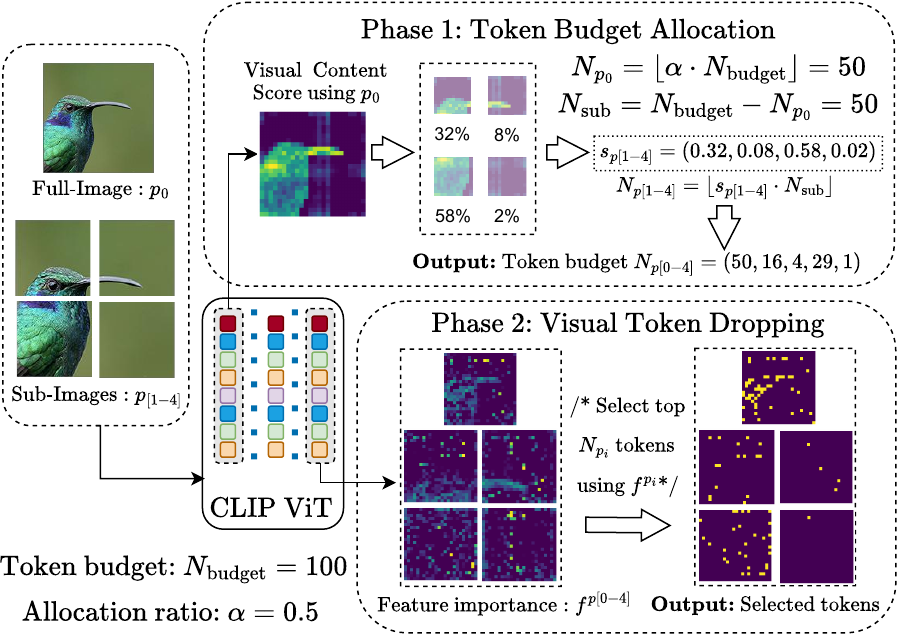}
    \caption{Design of \algoname{} for high-resolution VLMs to drop visual tokens before LLM. We first allocate token budgets for the full-image and sub-images and then select tokens with top feature importance within the allocated budget.
    }
    \label{fig:flowchart}
\end{figure}

To answer this question, we analyze the \texttt{CLS}-attention map across ViT layers and uncover two important findings. First, \texttt{CLS}-attention maps in \emph{initial layers} reveal the main content of the input image. The highlighted patches in these attention maps correspond to visually important parts of the image. As shown in \reffig{vit-cls-attn}, the attention maps of the first two layers highlight patches derived from \emph{the bird} while ignoring background areas with insignificant visual content. Second, attention maps in \emph{final layers} indicate the informative areas, i.e., patches containing more image features. In the last two layers, the highlighted patches are distributed across both the image content and the background. As the ViT processes the image, it encodes local features into corresponding patches in the initial layers. However, in the final layers, it learns the relationships between these local features and encodes them into a few background patches as global features~\cite{vit-need-register}. As a result, the highlighted areas in the \texttt{CLS}-attention of final layers prioritize patches that are more informative than the others~\cite{ia-red-vit}.

\begin{algorithm}[!t]
\footnotesize
    \caption{\algoname{}} 
    \label{alg:algorithm}
    \begin{algorithmic}[1]
        \STATE \textbf{Input}: $N_\mathrm{budget}$, $N_{\mathrm{ViT}}$, $\alpha$, $k$, $l_{\mathrm{init}}$, $l_{\mathrm{final}}$, 
        $H$, $T_{p_i}$,
        $\{a^{p_i}_{l,h}[j]\}$.
        \vspace{2mm}
        
        /* Phase 1: Token Budget Allocation */
        \\// 1-1. Compute the visual token budgets $N_{p_0}$ and $N_{\mathrm{sub}}$ for the full-image  and all the sub-images, respectively 
        \STATE $N_{{p_0}} \gets \lfloor \alpha \cdot N_{\mathrm{budget}} \rfloor$; 
        \STATE $N_{\mathrm{sub}} \gets N_{\mathrm{budget}} - N_{{p_0}}$; 

        // 1-2. Compute budget $N_{p_i}$ for each sub-image partition $p_i$ \\
        // Calculate visual content score $s_{p_i}$ for each sub-image partition $p_i$ using initial layer's \texttt{CLS}-attention of full-image $p_0$
        \STATE \textbf{For} each sub-image partition $p_i$ with $i = 1:k$ \textbf{do} 
                \STATE \hspace{1\algorithmicindent} $s_{p_i} \gets \sum_{j \in T_{p_i}} \sum_{h = 1}^{H} a^{p_0}_{l_{\mathrm{init}},h} [j]$; 

        \STATE \textbf{end For}
        \\ // Allocate budget $N_{p_i}$ for each sub-image partition $p_i$ 
        \STATE \textbf{For} each sub-image partition $p_i$ with $i = 1:k$ \textbf{do}     
            \STATE \hspace{1\algorithmicindent} $N_{p_i} \gets \lfloor N_{\mathrm{sub}} \cdot \frac{s_{p_i}}{\sum_{j=1}^k s_{p_j}} \rfloor$;
        \STATE \textbf{end For}

        \vspace{2mm}
        /* Phase 2: Visual Token Dropping */   
        \STATE \textbf{For} each image partition $p_i$ with $i = 0:k$ \textbf{do}
            \\\hspace{\algorithmicindent} // 2-1. Compute feature importance score $f^{p_i}[j]$ for each token $j$ using final layer's \texttt{CLS}-attention of image partition $p_i$
            \STATE \hspace{\algorithmicindent} \textbf{For} $j = 1:N_{\mathrm{ViT}}$ \textbf{do}
                \STATE \hspace{2\algorithmicindent} $f^{p_i}[j] \gets \sum_{h=1}^{H} a^{p_i}_{l_{\mathrm{final}},h}[j]$;
            \STATE \hspace{\algorithmicindent} \textbf{end For}
            \\\hspace{\algorithmicindent} // 2-2. Select important visual tokens within the budget
            \STATE \hspace{\algorithmicindent} Select top $N_{p_i}$ visual tokens with the highest $f^{p_i}[j]$;
            
        \STATE \textbf{end For}
    \end{algorithmic}
\end{algorithm}

\begin{table*}[!t]
    \centering
    \caption{Accuracy comparison between \algoname{} and the baselines. In all metrics, higher values indicate better performance. Here, VQA$^{\text{v2}}$, SQA, and VQA$^{\text{T}}$ stand for VQA-v2, ScienceQA, and TextVQA, respectively.
    } 
    \begin{tabular}{llccccccccc}
        \toprule
        \multirow{2}{*}{Model \& Method} & \multirow{2}{*}{Budget} & \multicolumn{2}{c}{{Visual QA}} & \multicolumn{3}{c}{{Transcription}} & \multicolumn{3}{c}{{Others}} \\
        \cmidrule(lr){3-4} \cmidrule(lr){5-7} \cmidrule(lr){8-10}
         &  & VQA$^{\text{v2}}$ & SQA & VQA$^{\text{T}}$ & DocVQA & OCRBench & MME & POPE & ChartQA \\
        \midrule
        LLaVA-Next-7B  & Full & 80.3 & 73.2 & 64.8 & 73.4 & 501 & 1519 & 87.6 & 54.8 \\ 
        \hspace{0.5cm} Spatial & 40\% & 77.7 & 68.0 & 57.0 & 58.9 & 369 & 1401 & 87.2 & 39.0 \\
        \hspace{0.5cm} PruMerge & 10\%* & 75.6 & 66.8 & 53.5 & 37.8 & 336 & 1393 & 85.0 & 28.8 \\ 
        \hspace{0.5cm} PruMerge+ & 55\%* & 78.0 & 68.2 & 54.4 & 44.6 & 365 & 1474 & 87.9 & 30.2 \\
        \rowcolor{LightOrange}
        \hspace{0.5cm} \algoname & 20\% & 77.5 & 73.4 & 61.4 & 60.8 & 475 & 1483 & 87.0 & 42.0 \\
        \rowcolor{DeeperOrange}
        \hspace{0.5cm} \algoname & 40\% & 78.8  & 73.8 & 63.6 & 68.7 & 488 & 1474 & 88.2 & 46.5 \\ 
        \midrule
        LLaVA-Next-13B & Full  & 80.9 & 73.6 & 66.9 & 77.5 & 508 & 1572 & 87.1 & 66.2 \\ 
        \hspace{0.5cm} Spatial & 40\% & 79.1 & 73.0 & 58.8 & 61.3 & 390 & 1529 & 87.2 & 42.6 \\
        \hspace{0.5cm} PruMerge & 10\%* & 74.1 & 69.2 & 54.4 & 45.9 & 381 & 1471 & 84.9 & 31.0 \\ 
        \hspace{0.5cm} PruMerge+ & 55\%* & 79.1 & 70.7 & 55.9 & 45.9 & 381 & 1480 & 87.5 & 31.0 \\ 
        \rowcolor{LightOrange}
        \hspace{0.5cm} \algoname & 20\% & 77.9 & 71.9 & 63.6 & 64.3 & 462 & 1545 & 86.7 & 48.9 \\
        \rowcolor{DeeperOrange}
        \hspace{0.5cm} \algoname & 40\% & 79.3 & 73.2 & 65.2 & 72.5 & 491 & 1570 & 87.7 & 53.7 \\ 
        \bottomrule
    \end{tabular}
    \label{tab:accuracy_results}
    \caption*{*Since PruMerge and PruMerge+ automatically determine the token budget, we report their average token usage.}
\end{table*}

\subsection{\algoname{} Design} \label{sec:design}

Inspired by the above findings on \texttt{CLS}-attention, we propose \algoname{}, an attention-guided token-dropping scheme comprising two phases: 1) \emph{Token Budget Allocation}, which determines the drop ratio for each image partition, given a total token budget; 2) \emph{Visual Token Dropping}, which selects the most informative visual tokens (and drops the rest) according to the drop ratio determined in Phase~1. The overall design is illustrated in Fig.~\ref{fig:flowchart} and detailed in Algorithm~\ref{alg:algorithm}.

\paragraph{Phase 1: Token Budget Allocation.} 
Let $
\{p_0,p_1,\dots,p_{k}\}$ 
denote the set of $(k+1)$ image partitions, where $p_0$ denotes the full-image and $p_i$ denotes the $i$-th sub-images. 
Let \(T_{p_i}\) denote the set of token indices of the full-image corresponding to sub-image \(p_i\). Each partition consists of $N_{\mathrm{ViT}}$ tokens (e.g., 576 for CLIP), totaling $N_{\mathrm{ViT}} \cdot (k+1)$ tokens before dropping. Given a token budget $N_{\mathrm{budget}}$, we allocate it across $(k+1)$ image partitions and use $N_{p_i}$ to denote the budget of image partition $p_i$. Consider the ViT consisting of $H$ heads and $L$ layers. Let $
\{a^{p_i}_{l,h}[j]\}_{j \in T_{p_i}}$ be the \texttt{CLS}-attention score at layer $l$ and head $h$ for partition $p_i$, and let $l_{\mathrm{init}}$ and $l_{\mathrm{final}}$ be the initial and final layers, respectively.

First, we allocate the budget $N_{\mathrm{budget}}$ between the full-image and a set of all sub-images using a budget \emph{allocation ratio} $\alpha \in [0, 1]$. The budget for the full-image is $N_{{p_0}} := \lfloor \alpha \cdot N_{\mathrm{budget}} \rfloor$, and the remaining budget, $N_{\mathrm{sub}} := N_{\mathrm{budget}} - N_{{p_0}}$, is allocated to the sub-images. 
Through experiments, we determine
the optimal value of $\alpha$ (see \refsec{ablation}).

Second, we distribute the remaining budget $N_{\mathrm{sub}}$ across $k$~sub-images. 
As discussed in~\refsec{vit_attn_guided_dropping}, the \texttt{CLS}-attention map of the initial layer on the full-image captures the distribution of visual contents in the image. That is, higher attention corresponds to regions with more visual content, indicating that these regions require a larger token budget (i.e., less dropping). To formalize this, we compute a \emph{visual content score} $s_{p_i}$ 
for each sub-image $p_i$ (excluding the full-image $p_0$) as follows:
\begin{align}
    s_{p_i} := \sum\nolimits_{j \in T_{p_i}} \sum\nolimits_{h = 1}^{H} a^{p_0}_{l_{\mathrm{init}}, h} [j], & \forall i \in \{1, 2, \cdots, k\}.
\end{align}
Specifically, for each sub-image $p_i$, we aggregate the \texttt{CLS}-attention across all corresponding tokens on the full-image (i.e., $T_{p_i}$) and across all $H$ heads of the initial layer ($l_{\mathrm{init}}=0$). Then, the budget for each sub-image is determined by its fraction of the total: $N_{p_i} := \lfloor N_{\mathrm{sub}} \cdot \frac{s_{p_i}}{\sum_{j=1}^k s_{p_j}} \rfloor$. This token budget guides the token dropping in the next phase.

\paragraph{Phase 2: Visual Token Dropping.} 
Our token-dropping scheme aims to retain the most informative visual tokens. To achieve this, we introduce a \emph{feature importance score} for each partition, denoted by \(
\{ f^{p_i}[j]\}_{j \in \{1, 2, \cdots, N_{\mathrm{ViT}}\}}
\). 
As observed in \refsec{vit_attn_guided_dropping}, the \texttt{CLS}-attention map of the final layer highlights the informative patches from both an image partition's subject and background areas. Moreover, different heads learn different features~\cite{interpreting-clip}. 
We thus compute the feature importance score \(f^{p_i}[j]\) for the \(j\)-th token in each partition \(p_i\) as follows:
\begin{equation}
    f^{p_i}[j] := \sum\nolimits_{h=1}^{H} a^{p_i}_{l_{\mathrm{final}},h}[j],
\end{equation}
for all \(i \in \{0, 1, \cdots, k\}\) and \(j \in \{1, 2, \cdots, N_{\mathrm{ViT}}\}\). 
Specifically, we add \texttt{CLS}-attention of the final layer (\(l_{\mathrm{final}} = 22\) across all heads. We make these design choices based on the experiments on the impact of different layers and head aggregation strategies (see \refsec{ablation}). 

Finally, we select the \(N_{p_i}\) number of tokens with the highest feature importance score \(f^{p_i}[j]\) and drop the rest of the tokens for each partition $p_i$. 
The selected visual tokens (along with text and system tokens) are then concatenated and fed into the subsequent LLM.

\section{Evaluation}\label{sec:evaluation}

We evaluate \algoname{} on LLaVA-Next~\cite{llava-next-technical}, LLaVA-v1.5~\cite{llava-main}, and ShareGPT4V~\cite{sharegpt}. For performance evaluation, we use an entry-level NVIDIA TESLA P40 (24~GB) GPU.

\paragraph{Downstream Tasks and Benchmarks.}
We used eight benchmarks from LMMS-EVAL~\cite{lmms-eval} evaluation framework across three different task types: 1) \emph{Visual Question Answering (VQA)} includes high-level object recognition benchmarks such as VQA-v2~\cite{vqav2} and ScienceQA~\cite{scienceqa}; 2) \emph{Transcription} focuses on fine-grained transcription tasks, including TextVQA~\cite{textvqa}, DocVQA~\cite{docvqa}, and OCRBench~\cite{orcbench}; and 3) \emph{Others} consists of MME~\cite{mme} for perception and cognition abilities, POPE~\cite{pope} for hallucination detection and ChartQA~\cite{chartqa} for spatial understanding.

\paragraph{Baselines.} We select PruMerge and PruMerge+~\cite{llava-prumerge} as our primary baselines because they utilize early-dropping mechanisms similar to ours. PruMerge uses spatial redundancy in visual tokens and performs pruning and merging on visual tokens, and PruMerge+, an enhanced version, additionally includes visual tokens through spatially uniform sampling to minimize accuracy losses. Since PruMerge and PruMerge+ are designed for LLaVA (a low-resolution VLM without dynamic partitioning), we apply their token-dropping strategy to each image partition. Additionally, we include spatial pooling as a simple baseline suggested in previous works~\cite{pooling}. While TokenCorrCompressor~\cite{flexattention} also supports early token dropping, they consider document understanding tasks only, and the code was not publicly available at the time of writing, rendering a direct comparison infeasible. Although FastV~\cite{fastv} and FlexAttention~\cite{flexattention} optimize VLMs through sparse attention in LLMs, they require processing excessive visual tokens in the earlier layers of LLMs, leading to inefficiencies in both latency and memory compared to early-dropping methods.

\subsection{Accuracy} \label{sec:accuracy_evaluation}

We evaluate the accuracy of \algoname{} on LLaVA-Next (7B and 13B) and compare it with the baselines for various tasks. The results are presented in Table~\ref{tab:accuracy_results}. To study the robustness of \algoname{}, we further evaluate low-resolution VLMs (with a single partition) such as LLaVA-1.5-7B and ShareGPT4V. The accuracy metrics reported in the table are the default metrics used for the corresponding tasks.

\paragraph{Accuracy vs. Token Reduction.} Evaluation results show that with a 20\% token budget (i.e., a maximum of 576 tokens), \algoname{} achieves nearly the same accuracy as full execution (i.e., a maximum of 2880 tokens) for VQA tasks. With a 40\% token budget (i.e., a maximum of 1152 tokens), it maintains comparable accuracy for fine-grained transcription tasks. Interestingly, for ScienceQA and POPE, we observe an increase in accuracy with fewer tokens. This suggests that in some cases, reducing the number of tokens to some extent may even improve accuracy.

\begin{table}[!t]
    \centering
    \caption{Inference efficiency:
    throughput, latency, and GPU memory usage across different batch sizes using LLaVA-Next-7B with \algoname{} under various token budgets.}
    \begin{tabular}{cccccc}
        \toprule
         Batch & \multirow{2}{*}{Budget} & Throughput & Latency & Memory \\
          Size & & (tokens/sec) & (sec) & (GB) \\
         \midrule
          & Full & 0.49 & 19.49 & 16.04 \\
        1 & 40\% & 1.40 & 6.91 & 14.33 \\
          & 20\% & 2.30 & 4.21 & 13.76 \\
         \midrule
          & Full & 0.66 & 44.37 & 21.84 \\
        2 & 40\% & 2.04 & 14.31 & 16.81 \\
          & 20\% & 3.68 & 7.89 & 15.07 \\
         \midrule
          & Full & -- & -- & OOM \\
        4 & 40\% & 2.22 & 26.38 & 20.36 \\
          & 20\% & 4.28 & 13.60 & 16.99 \\
        \bottomrule
    \end{tabular}
    \label{tab:inference}
\end{table}

\paragraph{Comparison with Baselines.} We observe a greater accuracy degradation across all tasks for both PruMerge and PruMerge+. While these methods can dynamically adjust the token budget to retain more visual information when necessary, they still fall short compared to \algoname{}, particularly in transcription tasks. On average, PruMerge and PruMerge+ use 10\% and 55\% of tokens, respectively, for transcription tasks (see Section~\ref{sec:cost_evaluation}). However, PruMerge+, even with 55\% of tokens (on average), has 11\% and 26\% lower accuracy than \algoname{}-20\% for TextVQA and DocVQA, respectively. Similarly, PruMerge with 10\% tokens, has 13\% and 37\% lower accuracy, respectively.

\subsection{Inference Efficiency} \label{sec:cost_evaluation}

\begin{figure}
    \centering
    \includegraphics[width=0.9\linewidth]{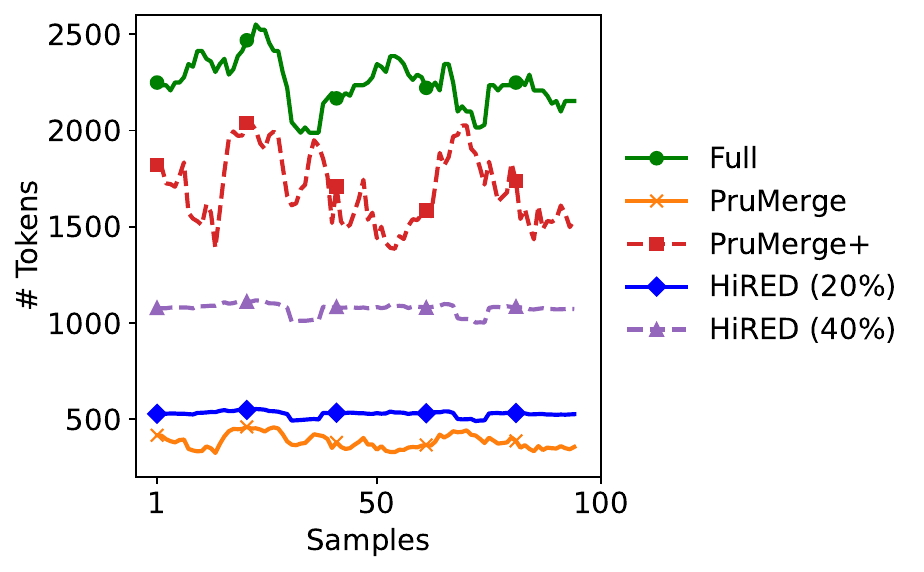}
    \caption{Number of visual tokens generated in 100 samples of TextVQA for Full, PruMerge, PruMerge+, and \algoname{}
    }
    \label{fig:tokens}
\end{figure}

To evaluate the inference efficiency of \algoname{}, we measure: 1)~the number of visual tokens; 2)~token generation throughput; 3)~time-to-first-token (TTFT) latency; 4)~GPU memory usage. The results are presented in Table~\ref{tab:inference}. A comparison of token usage with the baselines is illustrated in \reffig{tokens}.

\paragraph{Inference Efficiency of \algoname{}.}
Table~\ref{tab:inference} highlights the inference efficiency of \algoname{}. With a 20\% token budget, \algoname{} achieves a 4.7$\times$ increase in token generation throughput (2.30 vs. 0.49 tokens/sec) compared to the full execution (i.e., 100\% tokens). It also reduces the TTFT latency by 78\% (4.21 vs. 19.49 seconds), which is crucial for low-latency applications. Furthermore, \algoname{}-20\% reduces GPU memory usage by 14\% (13.76 vs. 16.04 GB). For larger batch sizes, it even shows higher gain. For instance, with a batch size of 4, full execution encounters out-of-memory (OOM) on the 24 GB GPU. In contrast, \algoname{}-20\% reduces memory usage by 30\% (16.99 GB) while maintaining the throughput and latency improvements.

\paragraph{Efficiency under Token Budget.}
\reffig{tokens} demonstrates that \algoname{}’s token usage in LLaVA-Next-7B consistently remains within the predefined resource constraints (e.g., 20\%). In contrast, full execution without dropping (Full), PruMerge, and PruMerge+ exhibit significant variations in the number of visual tokens across different samples in TextVQA. For Full, the variation arises from differences in partitioning based on the width-to-height ratio and resolution of image as well as the removal of some padding tokens. For PruMerge and PruMerge+, the variation stems from their adaptive nature, which allocates more tokens to images with higher visual information density. This fluctuation in the number of visual tokens directly affects computational costs, while \algoname{} enforces a strict token budget, making it well-suited 
in resource constraints.

\subsection{Ablation Study} \label{sec:ablation}

\begin{figure}[!t]
    \centering
    \includegraphics[width=0.48\columnwidth]{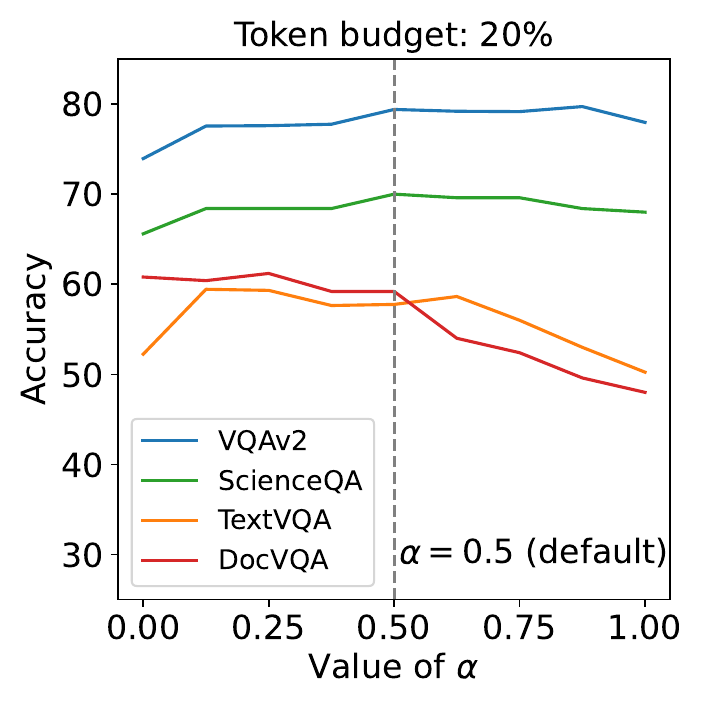}\hspace{2mm}
    \includegraphics[width=0.48\columnwidth]{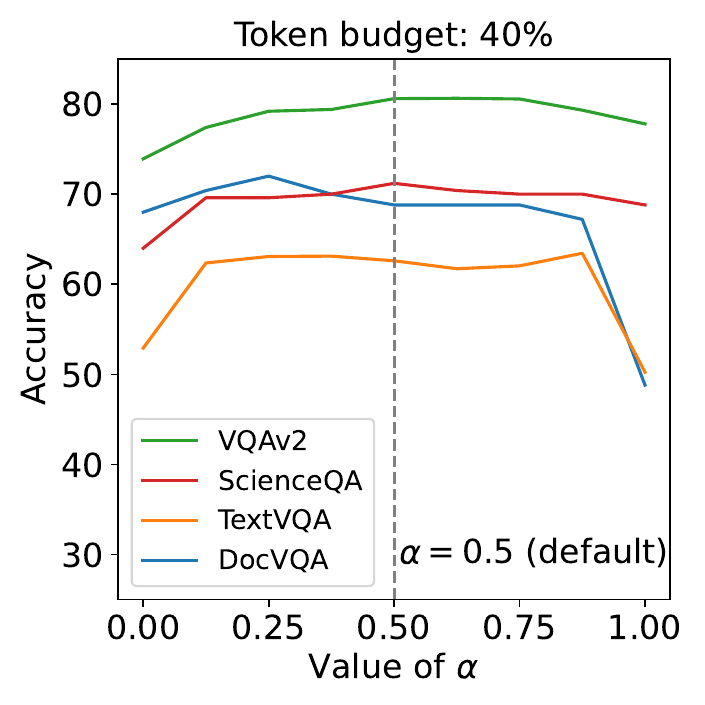}
    \caption{Choice of budget allocation ratio $\alpha$.
    }
    \label{fig:alpha}
\end{figure}

\begin{table}[!t]
    \centering
    \caption{Ablation study of ViT layer selection and head aggregation strategy in \algoname's token dropping algorithm.} 
    \setlength{\tabcolsep}{1.5mm}
    \begin{tabular}{lcccc}
        \toprule
         Choice & SQA & VQA$^{\text{v2}}$ & VQA$^{\text{T}}$ & DocVQA   \\ \toprule
         Distribute budget:\\
         \hspace{3mm} evenly & 67.6 & 73.3 & 52.2 & 48.8 \\
         \hspace{3mm} using layer 22 & 65.2 & 68.6 & 37.3 & 54.0 \\
         \hspace{3mm} \textbf{using layer 0} & \textbf{68.4} & \textbf{77.7} & \textbf{54.8} & \textbf{59.2} \\ \midrule
         Drop tokens:\\
         \hspace{3mm} using layer 0 & 65.2 & 68.6 & 37.3 & 54.0 \\
         \hspace{3mm} using layer 11 & 68.0 & 76.1 & 51.9 & 52.0 \\
         \hspace{3mm} \textbf{using layer 22}  & \textbf{68.4} & \textbf{77.7} & \textbf{54.8} & \textbf{59.2} \\ \midrule
         Aggregate heads:   \\ 
         \hspace{4mm}No agg & 67.2 & 76.6 & 50.5 & 52.4  \\ 
         \hspace{4mm}\textbf{Addition} & \textbf{68.4} & \textbf{77.7} & \textbf{54.8} & \textbf{59.2} \\ \bottomrule
    \end{tabular}
    \label{tab:design_choice}
\end{table}

We evaluate the key components of \algoname{} and the effectiveness of our design choices
. For this study, we select two VQA tasks (i.e., ScienceQA and VQA-v2) and two fine-grained transcription tasks (i.e., TextVQA and DocVQA).

\paragraph{Value of Budget Allocation Ratio \(\alpha\).}  \label{eval:budget_distribution_ablation}
Fig.~\ref{fig:alpha} shows that budget allocation ratio \(\alpha\) can impact the accuracy. 
Specifically, choosing \(\alpha = 0\) indicates no token budget allocated to the full-image, while \(\alpha = 1\) assigns the entire token budget to the full-image. The remaining token budget is distributed among the sub-images based on their importance. This result highlights that balancing the budget between the full-image and sub-images is crucial, and a balanced budget distribution (i.e., \(\alpha = 0.5\)) generally yields the highest accuracy. Therefore, we choose \(\alpha = 0.5\) as the default value for allocating the token budget between the full-image and sub-images.

\paragraph{Design Choices.} \label{eval:token_selection_ablation}  
We evaluate the design choices of \algoname{} in Table~\ref{tab:design_choice}. We use the \texttt{CLS}-attention from the initial ViT layer ($l_{\mathrm{init}}=0$
) to allocate the token budget. This choice is motivated by the stronger alignment of early-layer attentions with the visual content of the input image, compared to deeper layers like Layer~22. Even when the budget is distributed evenly across image partitions (e.g., dividing a budget of 100 among 5 partitions as 20 each), using Layer~0 achieves the best results. We determine the drop ratios using Layer~0 for token dropping but perform the actual dropping based on Layer~22. Dropping tokens using other layers (e.g., Layer~0 or Layer~11) leads to lower performance, as deeper layers in ViTs aggregate information into fewer, more informative tokens. Thus, using the final layer helps identify the most critical tokens. Furthermore, our head aggregation strategy, which combines attention scores through summation, achieves higher accuracy compared to no aggregation.

\paragraph{Low-Resolution VLMs.}  
We further evaluate \algoname{} for two low-resolution VLMs (i.e., models without dynamic partitioning) such as LLaVA-1.5-7B and ShareGPT4V-7B. This evaluation serves two purposes: 1) to demonstrate the robustness and wide applicability of \algoname{} across different VLMs; 2) to isolate the performance of the token-dropping strategy by excluding budget allocation, which does not apply to single-partition inputs. As shown in Table~\ref{tab:single_img_acc}, \algoname{} maintains accuracy close to full execution for LLaVA-1.5-7B and ShareGPT4V-7B, even with a limited token budget (40\% and 20\%). This result highlights the robustness and effectiveness of our token-dropping scheme.

\begin{table}[!t]
    \centering
    \caption{
    Accuracy comparison on low-resolution (i.e., single partition) using LLaVA-1.5-7B and ShareGPT4V-7B.
    }
    \setlength{\tabcolsep}{0.85mm}
    \begin{tabular}{lccccc}
        \toprule
        Model & Budget & SQA & VQA$^{\text{v2}}$ & VQA$^{\text{T}}$ & DocVQA \\ \midrule
        LLaVA-1.5-7B & Full & 69.5 & 76.6 & 46.1 & 28.1 \\
        \hspace{0.4cm} \algoname{} & 40\% & 67.2 & 79.0 & 47.0 & 29.4 \\
        \hspace{0.4cm} \algoname{} & 20\%  & 66.4 & 74.7 & 44.2 & 24.6  \\ \midrule
        ShareGPT4V-7B & Full & 68.4 & 80.6 & 50.7 & 26.76 \\
        \hspace{0.4cm} \algoname{} & 40\% & 67.2 & 79.0 & 50.0 & 25.97  \\
        \hspace{0.4cm} \algoname{} & 20\%  & 66.4 & 74.7 & 49.3 & 24.76 \\ \midrule
    \end{tabular}
    \label{tab:single_img_acc}
\end{table}

\section{Conclusion}

High-resolution VLMs significantly enhance multimodal capability by retaining detailed image information, but the excessive number of visual tokens poses significant challenges during inference. To address this challenge, we proposed \algoname{}, a plug-and-play token-dropping framework that allocates a fixed token budget across image partitions, prioritizes the most informative visual tokens, and drops the rest before LLM generation. Our evaluations demonstrate that \algoname{} substantially improves inference throughput, reduces latency, and lowers GPU memory consumption while maintaining competitive accuracy across diverse multimodal benchmarks. We believe \algoname{} provides a practical and scalable solution for deploying high-resolution VLMs in resource-constrained environments and offers a foundation for further optimization of multimodal inference systems.  

A limitation of \algoname{} is the potential loss of spatial information, which may impact tasks where spatial relationships are crucial, such as task understanding in ChartQA. This limitation arises as LLMs rely on positional encodings optimized for language modeling, which are less suited for visual tokens. One possible solution is to incorporate 2D positional encodings after token dropping to preserve spatial relationships, which we leave for future work.

\section*{Acknowledgments}

This work was supported in part by the National Science Foundation under Grants CNS-2315851 and 2106634, the Commonwealth Cyber Initiative, a Sony Faculty Innovation Award under AG3ZURVF, the Department for the Economy, Northern Ireland, under grant agreement USI-226, and Science Foundation Ireland under grant agreement 22/US/3848.

\bibliography{aaai25}

\begin{thebibliography}{46}
\providecommand{\natexlab}[1]{#1}

\bibitem[{Achiam et~al.(2024)Achiam, Adler, Agarwal, Ahmad, Akkaya, Aleman, Almeida, Altenschmidt, Altman, Anadkat et~al.}]{gpt4vtechnicalreport}
Achiam, J.; Adler, S.; Agarwal, S.; Ahmad, L.; Akkaya, I.; Aleman, F.~L.; Almeida, D.; Altenschmidt, J.; Altman, S.; Anadkat, S.; et~al. 2024.
\newblock GPT-4 Technical Report.
\newblock arXiv:2303.08774.

\bibitem[{Bai et~al.(2023)Bai, Bai, Yang, Wang, Tan, Wang, Lin, Zhou, and Zhou}]{qwen-vl-main}
Bai, J.; Bai, S.; Yang, S.; Wang, S.; Tan, S.; Wang, P.; Lin, J.; Zhou, C.; and Zhou, J. 2023.
\newblock Qwen-VL: A Versatile Vision-Language Model for Understanding, Localization, Text Reading, and Beyond.
\newblock arXiv:2308.12966.

\bibitem[{Cai et~al.(2024)Cai, Yang, Gao, and Lee}]{matryoshka-mm}
Cai, M.; Yang, J.; Gao, J.; and Lee, Y.~J. 2024.
\newblock Matryoshka Multimodal Models.
\newblock arXiv:2405.17430.

\bibitem[{Cao et~al.(2024)Cao, Ye, Li, Yu, Tang, Lu, and Chen}]{madtp}
Cao, J.; Ye, P.; Li, S.; Yu, C.; Tang, Y.; Lu, J.; and Chen, T. 2024.
\newblock MADTP: Multimodal Alignment-Guided Dynamic Token Pruning for Accelerating Vision-Language Transformer.
\newblock In \emph{Proceedings of the IEEE/CVF Conference on Computer Vision and Pattern Recognition (CVPR)}, 15710--15719.

\bibitem[{Cao, Paranjape, and Hajishirzi(2023)}]{pumer}
Cao, Q.; Paranjape, B.; and Hajishirzi, H. 2023.
\newblock {P}u{M}er: Pruning and Merging Tokens for Efficient Vision Language Models.
\newblock In \emph{Proceedings of the 61st Annual Meeting of the Association for Computational Linguistics (Volume 1: Long Papers)}, 12890--12903. Toronto, Canada: Association for Computational Linguistics (ACL).

\bibitem[{Cha et~al.(2024)Cha, Kang, Mun, and Roh}]{abstractor}
Cha, J.; Kang, W.; Mun, J.; and Roh, B. 2024.
\newblock Honeybee: Locality-enhanced projector for multimodal llm.
\newblock In \emph{Proceedings of the IEEE/CVF Conference on Computer Vision and Pattern Recognition}, 13817--13827.

\bibitem[{Chen et~al.(2025{\natexlab{a}})Chen, Li, Dong, Zhang, He, Wang, Zhao, and Lin}]{sharegpt}
Chen, L.; Li, J.; Dong, X.; Zhang, P.; He, C.; Wang, J.; Zhao, F.; and Lin, D. 2025{\natexlab{a}}.
\newblock Sharegpt4v: Improving large multi-modal models with better captions.
\newblock In \emph{European Conference on Computer Vision (ECCV)}, 370--387.

\bibitem[{Chen et~al.(2025{\natexlab{b}})Chen, Zhao, Liu, Bai, Lin, Zhou, and Chang}]{fastv}
Chen, L.; Zhao, H.; Liu, T.; Bai, S.; Lin, J.; Zhou, C.; and Chang, B. 2025{\natexlab{b}}.
\newblock An Image is Worth 1/2 Tokens After Layer 2: Plug-and-Play Inference Acceleration for Large Vision-Language Models.
\newblock In Leonardis, A.; Ricci, E.; Roth, S.; Russakovsky, O.; Sattler, T.; and Varol, G., eds., \emph{Computer Vision -- ECCV 2024}, 19--35. Cham: Springer Nature Switzerland.
\newblock ISBN 978-3-031-73004-7.

\bibitem[{Chen, Pekis, and Brown(2024)}]{llava-med-high-res}
Chen, Z.; Pekis, A.; and Brown, K. 2024.
\newblock Advancing High Resolution Vision-Language Models in Biomedicine.
\newblock arXiv:2406.09454.

\bibitem[{Chu et~al.(2023)Chu, Qiao, Lin, Xu, Yang, Hu, Wei, Zhang, Zhang, Wei, and Shen}]{mobile-vlm}
Chu, X.; Qiao, L.; Lin, X.; Xu, S.; Yang, Y.; Hu, Y.; Wei, F.; Zhang, X.; Zhang, B.; Wei, X.; and Shen, C. 2023.
\newblock MobileVLM : A Fast, Strong and Open Vision Language Assistant for Mobile Devices.
\newblock arXiv:2312.16886.

\bibitem[{Darcet et~al.(2024)Darcet, Oquab, Mairal, and Bojanowski}]{vit-need-register}
Darcet, T.; Oquab, M.; Mairal, J.; and Bojanowski, P. 2024.
\newblock Vision Transformers Need Registers.
\newblock In \emph{The Twelfth International Conference on Learning Representations}.

\bibitem[{Dettmers et~al.(2022)Dettmers, Lewis, Belkada, and Zettlemoyer}]{int8-quantization}
Dettmers, T.; Lewis, M.; Belkada, Y.; and Zettlemoyer, L. 2022.
\newblock GPT3.int8(): 8-bit Matrix Multiplication for Transformers at Scale.
\newblock In Koyejo, S.; Mohamed, S.; Agarwal, A.; Belgrave, D.; Cho, K.; and Oh, A., eds., \emph{Advances in Neural Information Processing Systems}, volume~35, 30318--30332. Curran Associates, Inc.

\bibitem[{Dong et~al.(2024)Dong, Zhang, Zang, Cao, Wang, Ouyang, Zhang, Duan, Zhang, Li, Yan, Gao, Chen, xinyue zhang, Li, Jingwen, Wang, Chen, He, ZHANG, Dai, Qiao, Lin, and Wang}]{internlm-xcomposer2}
Dong, X.; Zhang, P.; Zang, Y.; Cao, Y.; Wang, B.; Ouyang, L.; Zhang, S.; Duan, H.; Zhang, W.; Li, Y.; Yan, H.; Gao, Y.; Chen, Z.; xinyue zhang; Li, W.; Jingwen, L.; Wang, W.; Chen, K.; He, C.; ZHANG, X.; Dai, J.; Qiao, Y.; Lin, D.; and Wang, J. 2024.
\newblock Intern{LM}-{XC}omposer2-4{KHD}: A Pioneering Large Vision-Language Model Handling Resolutions from 336 Pixels to 4K {HD}.
\newblock In \emph{The Thirty-eighth Annual Conference on Neural Information Processing Systems}.

\bibitem[{Fu et~al.(2024)Fu, Chen, Shen, Qin, Zhang, Lin, Yang, Zheng, Li, Sun, Wu, and Ji}]{mme}
Fu, C.; Chen, P.; Shen, Y.; Qin, Y.; Zhang, M.; Lin, X.; Yang, J.; Zheng, X.; Li, K.; Sun, X.; Wu, Y.; and Ji, R. 2024.
\newblock MME: A Comprehensive Evaluation Benchmark for Multimodal Large Language Models.
\newblock arXiv:2306.13394.

\bibitem[{Gandelsman, Efros, and Steinhardt(2024)}]{interpreting-clip}
Gandelsman, Y.; Efros, A.~A.; and Steinhardt, J. 2024.
\newblock Interpreting {CLIP}'s Image Representation via Text-Based Decomposition.
\newblock In \emph{The Twelfth International Conference on Learning Representations}.

\bibitem[{Goyal et~al.(2017)Goyal, Khot, Summers-Stay, Batra, and Parikh}]{vqav2}
Goyal, Y.; Khot, T.; Summers-Stay, D.; Batra, D.; and Parikh, D. 2017.
\newblock Making the v in vqa matter: Elevating the role of image understanding in visual question answering.
\newblock In \emph{Proceedings of the IEEE/CVF Conference on Computer Vision and Pattern Recognition (CVPR)}, 6904--6913.

\bibitem[{Hu et~al.(2024)Hu, Xu, Ye, Yan, Zhang, Zhang, Zhang, Jin, Huang, and Zhou}]{docowl-1.5}
Hu, A.; Xu, H.; Ye, J.; Yan, M.; Zhang, L.; Zhang, B.; Zhang, J.; Jin, Q.; Huang, F.; and Zhou, J. 2024.
\newblock m{PLUG}-{D}oc{O}wl 1.5: Unified Structure Learning for {OCR}-free Document Understanding.
\newblock In Al-Onaizan, Y.; Bansal, M.; and Chen, Y.-N., eds., \emph{Findings of the Association for Computational Linguistics: EMNLP 2024}, 3096--3120. Miami, Florida, USA: Association for Computational Linguistics.

\bibitem[{Li et~al.(2025)Li, Chen, Cai, Chen, Hong, Chen, Shen, and Gan}]{flexattention}
Li, J.; Chen, D.; Cai, T.; Chen, P.; Hong, Y.; Chen, Z.; Shen, Y.; and Gan, C. 2025.
\newblock FlexAttention for Efficient High-Resolution Vision-Language Models.
\newblock In Leonardis, A.; Ricci, E.; Roth, S.; Russakovsky, O.; Sattler, T.; and Varol, G., eds., \emph{Computer Vision -- ECCV 2024}, 286--302. Cham: Springer Nature Switzerland.
\newblock ISBN 978-3-031-72698-9.

\bibitem[{Li et~al.(2023{\natexlab{a}})Li, Li, Savarese, and Hoi}]{blip2}
Li, J.; Li, D.; Savarese, S.; and Hoi, S. 2023{\natexlab{a}}.
\newblock Blip-2: Bootstrapping language-image pre-training with frozen image encoders and large language models.
\newblock In \emph{International conference on machine learning (ICML)}, 19730--19742. PMLR.

\bibitem[{Li et~al.(2023{\natexlab{b}})Li, Du, Zhou, Wang, Zhao, and Wen}]{pope}
Li, Y.; Du, Y.; Zhou, K.; Wang, J.; Zhao, X.; and Wen, J.-R. 2023{\natexlab{b}}.
\newblock Evaluating Object Hallucination in Large Vision-Language Models.
\newblock In Bouamor, H.; Pino, J.; and Bali, K., eds., \emph{Proceedings of the 2023 Conference on Empirical Methods in Natural Language Processing}, 292--305. Singapore: Association for Computational Linguistics.

\bibitem[{Li et~al.(2024)Li, Yang, Liu, Ma, Zhang, Yang, Sun, Liu, and Bai}]{monkey-main}
Li, Z.; Yang, B.; Liu, Q.; Ma, Z.; Zhang, S.; Yang, J.; Sun, Y.; Liu, Y.; and Bai, X. 2024.
\newblock Monkey: Image resolution and text label are important things for large multi-modal models.
\newblock In \emph{Proceedings of the IEEE/CVF Conference on Computer Vision and Pattern Recognition}, 26763--26773.

\bibitem[{Lin et~al.(2023)Lin, Liu, Zhang, Gao, Qiu, Xiao, Qiu, Lin, Shao, Chen, Han, Huang, Zhang, He, Li, and Qiao}]{sphinx-main}
Lin, Z.; Liu, C.; Zhang, R.; Gao, P.; Qiu, L.; Xiao, H.; Qiu, H.; Lin, C.; Shao, W.; Chen, K.; Han, J.; Huang, S.; Zhang, Y.; He, X.; Li, H.; and Qiao, Y. 2023.
\newblock SPHINX: The Joint Mixing of Weights, Tasks, and Visual Embeddings for Multi-modal Large Language Models.
\newblock arXiv:2311.07575.

\bibitem[{Liu et~al.(2024{\natexlab{a}})Liu, Li, Li, Li, Zhang, Shen, and Lee}]{llava-next-technical}
Liu, H.; Li, C.; Li, Y.; Li, B.; Zhang, Y.; Shen, S.; and Lee, Y.~J. 2024{\natexlab{a}}.
\newblock LLaVA-NeXT: Improved reasoning, OCR, and world knowledge.

\bibitem[{Liu et~al.(2023)Liu, Li, Wu, and Lee}]{llava-main}
Liu, H.; Li, C.; Wu, Q.; and Lee, Y.~J. 2023.
\newblock Visual Instruction Tuning.
\newblock In \emph{NeurIPS}.

\bibitem[{Liu et~al.(2024{\natexlab{b}})Liu, Li, Huang, Yang, Yu, Li, Yin, Liu, Jin, and Bai}]{orcbench}
Liu, Y.; Li, Z.; Huang, M.; Yang, B.; Yu, W.; Li, C.; Yin, X.-C.; Liu, C.-L.; Jin, L.; and Bai, X. 2024{\natexlab{b}}.
\newblock OCRBench: on the hidden mystery of OCR in large multimodal models.
\newblock \emph{Science China Information Sciences}, 67(12).

\bibitem[{Liu et~al.(2024{\natexlab{c}})Liu, Yang, Liu, Li, Ma, Zhang, and Bai}]{text-monkey-main}
Liu, Y.; Yang, B.; Liu, Q.; Li, Z.; Ma, Z.; Zhang, S.; and Bai, X. 2024{\natexlab{c}}.
\newblock TextMonkey: An OCR-Free Large Multimodal Model for Understanding Document.
\newblock arXiv:2403.04473.

\bibitem[{Lu et~al.(2022)Lu, Mishra, Xia, Qiu, Chang, Zhu, Tafjord, Clark, and Kalyan}]{scienceqa}
Lu, P.; Mishra, S.; Xia, T.; Qiu, L.; Chang, K.-W.; Zhu, S.-C.; Tafjord, O.; Clark, P.; and Kalyan, A. 2022.
\newblock Learn to Explain: Multimodal Reasoning via Thought Chains for Science Question Answering.
\newblock In Oh, A.~H.; Agarwal, A.; Belgrave, D.; and Cho, K., eds., \emph{Advances in Neural Information Processing Systems}.

\bibitem[{Marin et~al.(2023)Marin, Chang, Ranjan, Prabhu, Rastegari, and Tuzel}]{pooling}
Marin, D.; Chang, J.-H.~R.; Ranjan, A.; Prabhu, A.; Rastegari, M.; and Tuzel, O. 2023.
\newblock Token pooling in vision transformers for image classification.
\newblock In \emph{Proceedings of the IEEE/CVF Winter Conference on Applications of Computer Vision}, 12--21.

\bibitem[{Masry et~al.(2022)Masry, Do, Tan, Joty, and Hoque}]{chartqa}
Masry, A.; Do, X.~L.; Tan, J.~Q.; Joty, S.; and Hoque, E. 2022.
\newblock {C}hart{QA}: A Benchmark for Question Answering about Charts with Visual and Logical Reasoning.
\newblock In Muresan, S.; Nakov, P.; and Villavicencio, A., eds., \emph{Findings of the Association for Computational Linguistics: ACL 2022}, 2263--2279. Dublin, Ireland: Association for Computational Linguistics.

\bibitem[{Mathew, Karatzas, and Jawahar(2021)}]{docvqa}
Mathew, M.; Karatzas, D.; and Jawahar, C. 2021.
\newblock Docvqa: A dataset for vqa on document images.
\newblock In \emph{Proceedings of the IEEE/CVF Conference on Computer Vision and Pattern Recognition (CVPR)}, 2200--2209.

\bibitem[{McKinzie et~al.(2025)McKinzie, Gan, Fauconnier, Dodge, Zhang, Dufter, Shah, Du, Peng, Belyi et~al.}]{mm1-apple-main}
McKinzie, B.; Gan, Z.; Fauconnier, J.-P.; Dodge, S.; Zhang, B.; Dufter, P.; Shah, D.; Du, X.; Peng, F.; Belyi, A.; et~al. 2025.
\newblock MM1: methods, analysis and insights from multimodal LLM pre-training.
\newblock In \emph{European Conference on Computer Vision}, 304--323. Springer.

\bibitem[{Pan et~al.(2021)Pan, Panda, Jiang, Wang, Feris, and Oliva}]{ia-red-vit}
Pan, B.; Panda, R.; Jiang, Y.; Wang, Z.; Feris, R.; and Oliva, A. 2021.
\newblock IA-RED\^{}2: Interpretability-Aware Redundancy Reduction for Vision Transformers.
\newblock In Ranzato, M.; Beygelzimer, A.; Dauphin, Y.; Liang, P.; and Vaughan, J.~W., eds., \emph{Advances in Neural Information Processing Systems}, volume~34, 24898--24911. Curran Associates, Inc.

\bibitem[{Radford et~al.(2021)Radford, Kim, Hallacy, Ramesh, Goh, Agarwal, Sastry, Askell, Mishkin, Clark et~al.}]{clipvit}
Radford, A.; Kim, J.~W.; Hallacy, C.; Ramesh, A.; Goh, G.; Agarwal, S.; Sastry, G.; Askell, A.; Mishkin, P.; Clark, J.; et~al. 2021.
\newblock Learning transferable visual models from natural language supervision.
\newblock In \emph{International conference on machine learning}, 8748--8763. PMLR.

\bibitem[{Rao et~al.(2021)Rao, Zhao, Liu, Lu, Zhou, and Hsieh}]{dynamicvit}
Rao, Y.; Zhao, W.; Liu, B.; Lu, J.; Zhou, J.; and Hsieh, C.-J. 2021.
\newblock Dynamicvit: Efficient vision transformers with dynamic token sparsification.
\newblock \emph{Advances in neural information processing systems}, 34: 13937--13949.

\bibitem[{Reid et~al.(2024)Reid, Savinov, Teplyashin, Lepikhin, Lillicrap, Alayrac, Soricut, Lazaridou, Firat, Schrittwieser et~al.}]{gemini-1.5-pro}
Reid, M.; Savinov, N.; Teplyashin, D.; Lepikhin, D.; Lillicrap, T.; Alayrac, J.-b.; Soricut, R.; Lazaridou, A.; Firat, O.; Schrittwieser, J.; et~al. 2024.
\newblock Gemini 1.5: Unlocking multimodal understanding across millions of tokens of context.
\newblock arXiv:2403.05530.

\bibitem[{Shang et~al.(2024)Shang, Cai, Xu, Lee, and Yan}]{llava-prumerge}
Shang, Y.; Cai, M.; Xu, B.; Lee, Y.~J.; and Yan, Y. 2024.
\newblock LLaVA-PruMerge: Adaptive Token Reduction for Efficient Large Multimodal Models.
\newblock arXiv:2403.15388.

\bibitem[{Shi et~al.(2024)Shi, Tao, Rao, Yang, Yuan, and Wang}]{crossget}
Shi, D.; Tao, C.; Rao, A.; Yang, Z.; Yuan, C.; and Wang, J. 2024.
\newblock {C}ross{GET}: Cross-Guided Ensemble of Tokens for Accelerating Vision-Language Transformers.
\newblock In Salakhutdinov, R.; Kolter, Z.; Heller, K.; Weller, A.; Oliver, N.; Scarlett, J.; and Berkenkamp, F., eds., \emph{Proceedings of the 41st International Conference on Machine Learning (ICML)}, volume 235 of \emph{Proceedings of Machine Learning Research}, 44960--44990. PMLR.

\bibitem[{Singh et~al.(2019)Singh, Natarajan, Shah, Jiang, Chen, Batra, Parikh, and Rohrbach}]{textvqa}
Singh, A.; Natarajan, V.; Shah, M.; Jiang, Y.; Chen, X.; Batra, D.; Parikh, D.; and Rohrbach, M. 2019.
\newblock Towards VQA Models That Can Read.
\newblock In \emph{Proceedings of the IEEE/CVF Conference on Computer Vision and Pattern Recognition (CVPR)}.

\bibitem[{Sun et~al.(2024)Sun, Liu, Bair, and Kolter}]{pruning-wanda}
Sun, M.; Liu, Z.; Bair, A.; and Kolter, J.~Z. 2024.
\newblock A Simple and Effective Pruning Approach for Large Language Models.
\newblock In \emph{The Twelfth International Conference on Learning Representations}.

\bibitem[{Tao et~al.(2024)Tao, Chen, Jin, Bai, Zhao, and Lou}]{evit}
Tao, Z.; Chen, X.; Jin, Z.; Bai, X.; Zhao, H.; and Lou, Y. 2024.
\newblock {EVIT}: Event-Oriented Instruction Tuning for Event Reasoning.
\newblock In Ku, L.-W.; Martins, A.; and Srikumar, V., eds., \emph{Findings of the Association for Computational Linguistics: ACL 2024}, 8966--8979. Bangkok, Thailand: Association for Computational Linguistics.

\bibitem[{Vaswani(2017)}]{transformers}
Vaswani, A. 2017.
\newblock Attention is all you need.
\newblock \emph{Advances in Neural Information Processing Systems}.

\bibitem[{Zhang et~al.(2024{\natexlab{a}})Zhang, Yu, Dong, Li, Su, Chu, and Yu}]{survey-on-vlm-mm-llms}
Zhang, D.; Yu, Y.; Dong, J.; Li, C.; Su, D.; Chu, C.; and Yu, D. 2024{\natexlab{a}}.
\newblock {MM}-{LLM}s: Recent Advances in {M}ulti{M}odal Large Language Models.
\newblock In Ku, L.-W.; Martins, A.; and Srikumar, V., eds., \emph{Findings of the Association for Computational Linguistics: ACL 2024}, 12401--12430. Bangkok, Thailand: Association for Computational Linguistics.

\bibitem[{Zhang et~al.(2024{\natexlab{b}})Zhang, Li, Zhang, Pu, Cahyono, Hu, Liu, Zhang, Yang, Li, and Liu}]{lmms-eval}
Zhang, K.; Li, B.; Zhang, P.; Pu, F.; Cahyono, J.~A.; Hu, K.; Liu, S.; Zhang, Y.; Yang, J.; Li, C.; and Liu, Z. 2024{\natexlab{b}}.
\newblock LMMs-Eval: Reality Check on the Evaluation of Large Multimodal Models.
\newblock arXiv:2407.12772.

\bibitem[{Zhang et~al.(2024{\natexlab{c}})Zhang, Lyu, Shao, Chen, Guan, and Nie}]{token-corr-compress}
Zhang, R.; Lyu, Y.; Shao, R.; Chen, G.; Guan, W.; and Nie, L. 2024{\natexlab{c}}.
\newblock Token-level Correlation-guided Compression for Efficient Multimodal Document Understanding.
\newblock arXiv:2407.14439.

\bibitem[{Zhou et~al.(2024)Zhou, Hu, Weng, Jia, Luo, Liu, Wu, and Huang}]{tiny-llava}
Zhou, B.; Hu, Y.; Weng, X.; Jia, J.; Luo, J.; Liu, X.; Wu, J.; and Huang, L. 2024.
\newblock TinyLLaVA: A Framework of Small-scale Large Multimodal Models.
\newblock arXiv:2402.14289.

\bibitem[{Zhu et~al.(2024)Zhu, Zhu, Liu, Xu, and Peng}]{llava-phi}
Zhu, Y.; Zhu, M.; Liu, N.; Xu, Z.; and Peng, Y. 2024.
\newblock LLaVA-Phi: Efficient Multi-Modal Assistant with Small Language Model.
\newblock In \emph{Proceedings of the 1st International Workshop on Efficient Multimedia Computing under Limited}, EMCLR'24, 18–22. New York, NY, USA: Association for Computing Machinery.
\newblock ISBN 9798400711909.

\end{thebibliography}

\end{document}